\documentclass[letterpaper, 10 pt, conference]{ieeeconf}  

\IEEEoverridecommandlockouts                              

\usepackage{times}
\usepackage{epsfig}
\usepackage{graphicx}
\usepackage{amssymb}
\usepackage[pagebackref=true,breaklinks=true,letterpaper=true,colorlinks,bookmarks=false]{hyperref}
\usepackage{gensymb}
\usepackage{subfig}
\usepackage{algorithm}  
\usepackage[noend]{algpseudocode}  
\usepackage{amsmath}
\usepackage{array}
\usepackage{mars}

\title{\LARGE \bf
Mapping with Reflection - Detection and \\Utilization of Reflection in 3D Lidar Scans
}

\author{Xiting Zhao$^{1}$, Zhijie Yang$^{1}$ and S\"oren Schwertfeger$^{1}$
\thanks{$^{1}$All authors are with the School of Information Science and Technology, 
ShanghaiTech University, China.
        {\tt\small [zhaoxt, yangzhj, soerensch]@shanghaitech.edu.cn}}%
}

\begin{document}

%
%


\marsPublishedIn{Accepted for:} 		

\marsVenue{IEEE International Conference on Safety, Security, and Rescue Robotics (SSRR) 2020}

\marsYear{2020}

\marsPlainAutors{Xiting Zhao, Zhijie Yang and S\"oren Schwertfeger}


\marsMakeCitation{Mapping with Reflection - Detection and Utilization of Reflection in 3D Lidar Scans}{IEEE Press}


\marsIEEE{}


\makeMARStitle

%
%

\maketitle
\thispagestyle{empty}
\pagestyle{empty}




\addtolength{\belowcaptionskip}{-1pt}

\begin{abstract}
This paper presents a method to detect reflection of 3D light detection and ranging (Lidar) scans and uses it to classify the points and also map objects outside the line of sight. Our software uses several approaches to analyze the point cloud, including intensity peak detection, dual return detection, plane fitting, and finding the boundaries. These approaches can classify the point cloud and detect the reflection in it. By mirroring the reflection points on the detected window pane and adding classification labels on the points, we can improve the map quality in a Simultaneous Localization and Mapping (SLAM) framework. Experiments using real scan data and ground truth data showcase the effectiveness of our method. 
\end{abstract}

\section{Introduction}
\label{chap:introduction}
\begin{figure}[htbp]
\vspace{-0.3cm}
\centering
\includegraphics[width=7.5cm,height=3cm]{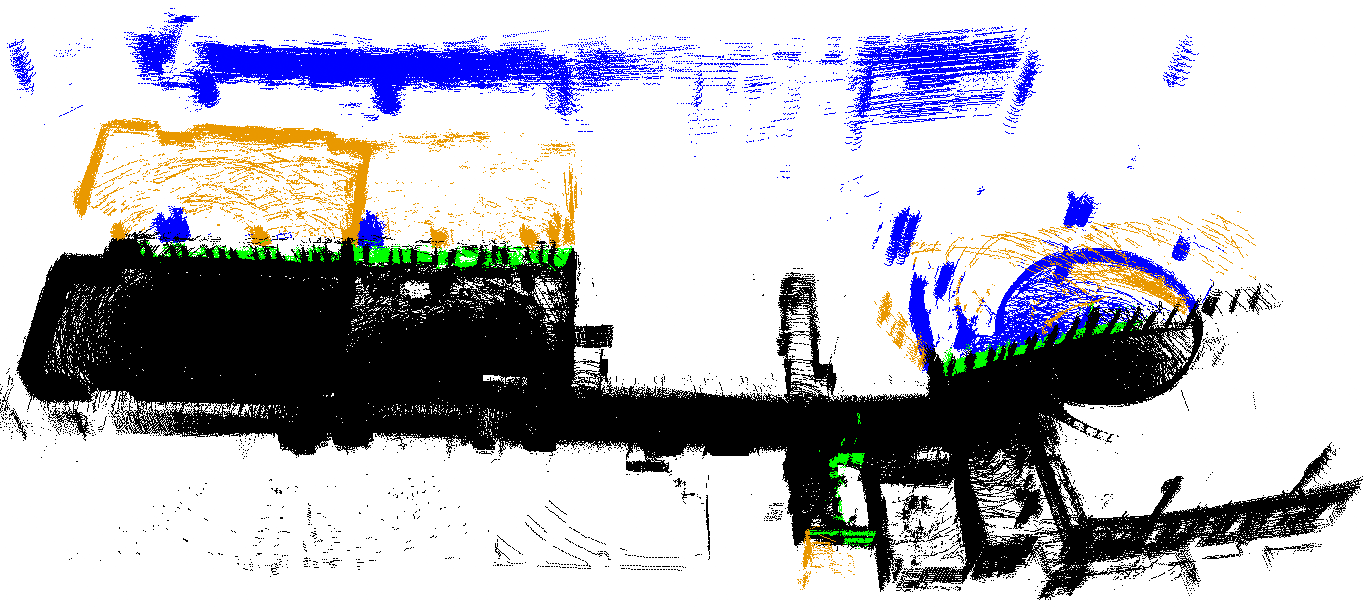}
\caption{Orange parts are the error in SLAM}
\label{reflectioninslam}
\vspace{-0.3cm}
\end{figure}

Simultaneous Localization And Mapping (SLAM) is quite often used by robots of industrial, service and rescue applications, to name a few. To get a better result of SLAM, light detection and ranging (Lidar) is used for its high accuracy, long detection range, and high stability\cite{cadena2016past}. However, reflective materials like glass and mirrors will cause problems when using Lidar. It can not only cause the sensor to report the wrong range data to the reflecting obstacle, and thus potentially causing a collision, but can also provide wrong reflected points, which will cause errors in the maps generated by a SLAM algorithm. For example, Fig. \ref{reflectioninslam} shows a SLAM result in a scene full of glass. The orange points are reflected by the window, black ones are from the concrete objects, blue ones are from the same source as black ones but passed through the glass, and the green ones are the reflective planes detected by our method.
As a result, the important fact that the detection and removal of the reflection can improve the map quality of SLAM and the safety of autonomous robots is obvious. Those reflected points will significantly degrade the map quality. In navigation application, wrong reflected points may cause problems in path planning. 
 The detection of windows and mirrors is also important for semantic segmentation of 3D point clouds\cite{ali2008robust,he2019furniturefree}.

In this paper, we propose a point cloud reflection detection method, based on off-the-shelf dual return Lidar, that has the following contributions: 1. The rejection of wrong reflected points and reflect back to correct position; 2. Points from reflective surfaces marked and treated as obstacles in the map; 3. Keeps points from real objects behind the window. 

The paper is organized as follows. Section \ref{chap:introduction} introduces the background and related work. Section \ref{chap:modeling} introduces the sensor and reflection modeling. In Section \ref{chap:methods}, we present methods to detect and remove the reflection and build the map. Experiments, results, and discussion are presented in Section \ref{chap:results}. Finally, conclusions are given in Section \ref{chap:conclusions}.

\subsection{Background}
\label{background}
The experiment and data collection is done by the MARS Jackal Mapper, a fully hardware-level synchronized mapping robot platform for 3D mapping and SLAM build by ShanghaiTech Mobile Autonomous Robotic Systems Lab (MARS Lab)\cite{CHEN2020103559}.  
This is a robot with powerful computation, RGB cameras, IMUs, robot odometry and two Velodyne HDL-32E Lidars (one installed vertically and the other horizontally), that is fully calibrated\cite{chen2019calibration}. For reflection detection we only use data from the horizontally scanning Velodyne.

\begin{figure}[htbp]
\centering
\includegraphics[width=8.5cm,height=2.5cm]{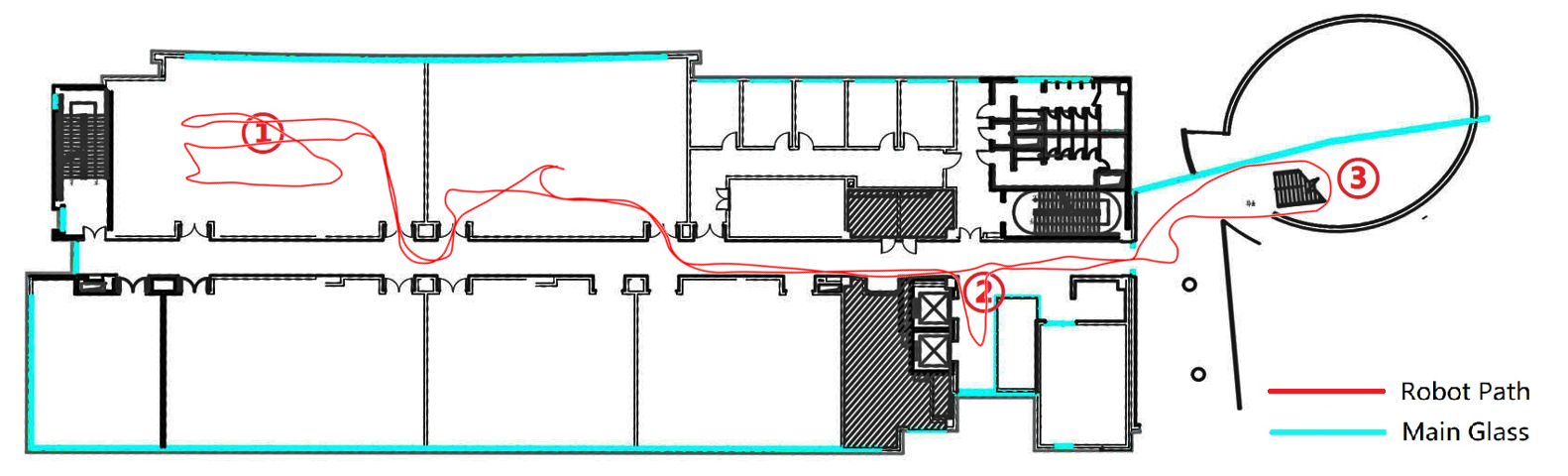}
\caption{Robot Path in STAR Center}
\label{starcenterpath}
\vspace{-0.5cm}
\end{figure}

We record the test and experimental data by driving a robot in the second floor of the STAR Center (ShangahiTech Automation and Robotics Center). The map of the STAR Center and the path of robot are shown in Fig. \ref{starcenterpath}. The red line approximates the robot path, blue lines show the main glass planes, while some small windows and glass are not marked. In the data, we have collected different kinds of reflection scenes, including windows, glass railing, glass door and floor-to-ceiling windows. These are common situations which usually cause problems for indoor SLAM. We will use this data to test our detection method and evaluate the result.

\subsection{Related Work}
\label{relatedworks}
There are already a number of papers on reflection detection in Lidar. Most of them are using 2D Lidars with full wave analysis. Others are about post-processing of data acquired by a 3D laser scanner with a camera. However, most of the commodity 3D Lidar scanners, such as the ones from Velodyne and Robosense, are only with two returns at most --- the strongest and the latest, and only little work has been done about reflection detection such kind of Lidar scanners.


Yang \textit{et al.} used mirror symmetry in 2D Lidar to resolve mirror reflection\cite{yang2011solving}. They firstly use a distance-based criterion to determine gaps in a laser scan. Then they use a Gaussian model to predict a potential mirror. After that, they use the Euclidean distance function to calculate the likelihood of the mirror for verification. Finally, they use ICP to match the reflected points and find the mirror.

Koch \textit{et al.} have researched on 2D multi-echo Lidar. They used a Hokuyo 30LX-EW Lidar, which records up to three echoes of the returning light wave, also with distances and intensities. They first use the sensor and integrate the pre-filter and post-filter to TSD slam to detect the reflection in 2D SLAM\cite{koch2016detection}. They also use different reflective intensity values to detect the transparent and specular reflective material, and improve their 2D slam\cite{koch2017identification}. For 3D mapping they rotated the 2D Lidar by mounting it on a motor\cite{koch2017detection}. 

Wang \textit{et al.} used a Velodyne 3D Lidar to detect windows while driving outdoors\cite{wang2011window}. They clustered the Lidar points and then detects the facades of the buildings. After that they calculated the surface normal by using PCA to detect potential windows. Finally, they projected potential window points to localize the windows.

Yun \textit{et al.} purpose a method to remove the reflection for large scale 3D point clouds\cite{yun2018reflection}. They find glass points on the unit sphere according to the number of echo pulses. Then they estimate the reliability of detected glass point and reflection symmetry to remove the virtual point.

Ali \textit{et al.} use the variability of laser scans of windows plus RGB images to find windows in facades of windows\cite{ali2008robust}.

Other methods use different sensors, such as high-speed time-of-flight sensors\cite{Velten2012RecoveringTS}, or additional sensors, such as ultrasonic sensors, in addition to cameras or Lidar for a consistent and accurate reflective plane detection in the cases that visual systems tend to fail. A recent research suggests that it is also possible to perform accurate glass detection with neural network using visual data\cite{Mei_2020_CVPR}.

\section{Sensor Modeling}
\label{chap:modeling}

\subsection{Lidar Sensor}
Light detection and ranging (Lidar) is using light detection to obtain the range of obstacles. The Lidar sensor we use is  Velodyne HDL-32E, a widely used Lidar in SLAM and automatic driving. The sensor has three different modes of processing the laser beam pulses: dual return mode, strongest return mode, and last return mode. Upon receiving the light from the single laser beam in one direction, the sensor analyzes the strongest and the last return and calculates their distance and intensity. A sensor in dual return mode will return both the strongest and the latest. If the strongest return is the same as last return, the second-strongest return will returned as the strongest. A point without sufficient intensity will be ignored.

\subsection{Reflection Model of Different Material}
Different material have different reflectivities and optical properties. Normal materials like wood, walls, or clothes mainly have diffuse reflection to the laser light, with little absorption and specular reflection, which is ideal for Lidars. Reflective materials will reflect the incident laser light like mirrors or glass do. Glass mainly has specular reflection and transmittance with a little diffuse reflection on laser light. Using these optical properties we can build a reflection model about how laser light interacts with the glass. 

In \cite{koch2017identification} work on lasers hitting different material has been performed. Research on light behavior of hitting the glass at different angles is presented in \cite{foster2013visagge} and \cite{kim2016localization}. Fig. \ref{reflection} shows the reflection model of a laser beam interacting with glass. When the laser beam hits the glass, there may be three different results of laser beam return. When the laser beam hits the glass almost perpendicularly, the intensity received peaks. When the angle of incidence decreases, the intensity drops quickly. As shown in Fig. \ref{degreeintensity}, if the angle of incidence decreases to a certain degree (also related to the distance), the return intensity will become too low to detect. As glass has transmittance, some of the light can pass through it. If there is something behind the glass, it will diffusely reflect the light and the return light can pass through the glass again to the receiver. So a Lidar can estimate the distance of things behind the glass with weakened intensity due to twice passing through the window. Glass also has specular reflection, so some of the light is reflected by the glass at the angle of incidence. If specularly reflected light hits something in front of the glass, it will reflect from the reverse path to the receiver, and thus the Lidar may get a reflected reading from an object with decreased intensity. The sensor, unaware of the reflection, will report the reflected point to be located behind the glass.

\newsavebox{\measurebox}
\begin{figure}
\centering
\sbox{\measurebox}{%
  \hspace{-0.2cm}
  \begin{minipage}[b]{4.3cm}
    \subfloat[Reflection model of glass]{\includegraphics[width=4.3cm,height=3.2cm]{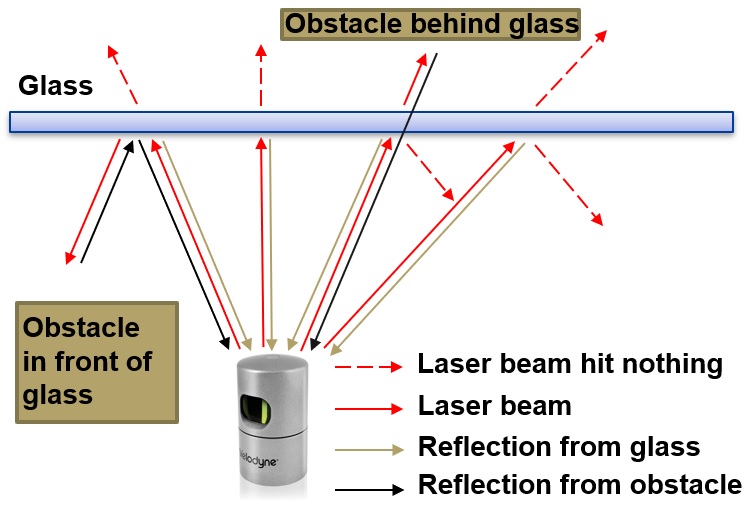}\label{reflection}}
  \vfill
  \subfloat[Degree of Reflection]{\includegraphics[width=4.3cm,height=1.3cm]{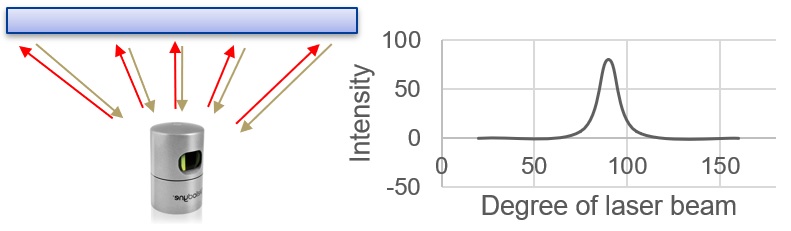}\label{degreeintensity}}
  \end{minipage}}
\usebox{\measurebox}\qquad
  \hspace{-0.6cm}
\begin{minipage}[b][\ht\measurebox][s]{4.2cm}
\centering
  \subfloat[Intensity of Reflection]{\includegraphics[width=4.2cm,height=5.4cm]{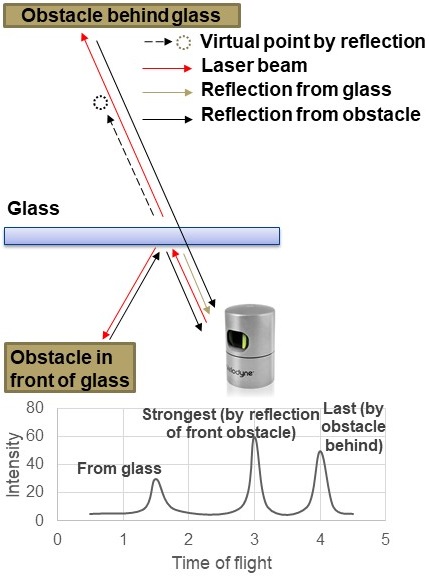}\label{reflectionintensity}}
\end{minipage}
\caption{Laser Reflection Model}
\vspace{-0.5cm}
\end{figure}

Fig. \ref{reflectionintensity} shows an example for the dual return mode of the Velodyne 3D Lidar scanner. In this case the sensor will measure three peaks. The first return is from the glass, because it is the nearest and thus the time of flight is the shortest. The second return is from the obstacle in front of the glass, and in this case it is the strongest. The last return is from the light that passes through the glass and is reflected by the obstacle behind glass. It is last in this example, because the path to the obstacle behind glass is longer than the path to the obstacle in front of the glass. So in this example, the strongest return will give a reflected point, the last will be the object behind the glass, while the glass itself is ignored because it is neither the strongest nor the last return.

From this analysis, we can conclude that: 1. The return from glass must be the nearest point of either the strongest or the last point. Moreover if there is more than one return, the return from the glass can only be the strongest point but not the last point. 2. The intensity returned from the glass will be the strongest when the incident laser beam is perpendicular to the glass and lower as the angle decreases. 3. Ignoring special conditions such as fog or smoke, if the strongest and last point differs, there must be a piece of glass in this direction (except for a few points that hit on the edges of obstacles partially, which may lead to different strongest and last points even without any reflective surfaces).

In addition to Velodyne, the data from scanners by other manufacturers, such as Robosense, also suits the analysis above.

We identified two phenomenons for being useful for reflection detection, which we will motivate below by analyzing example scans.

\subsection{Intensity Peak Analysis}
As mentioned before, if there is more than one return, the return from glass will be only possible to show in the strongest point but not the last point. So for this method we only use the strongest point cloud to analysis.

\begin{figure}[htbp]
\centering
\subfloat[Floor-to-ceiling glass]{
\includegraphics[width=4cm,height=2cm]{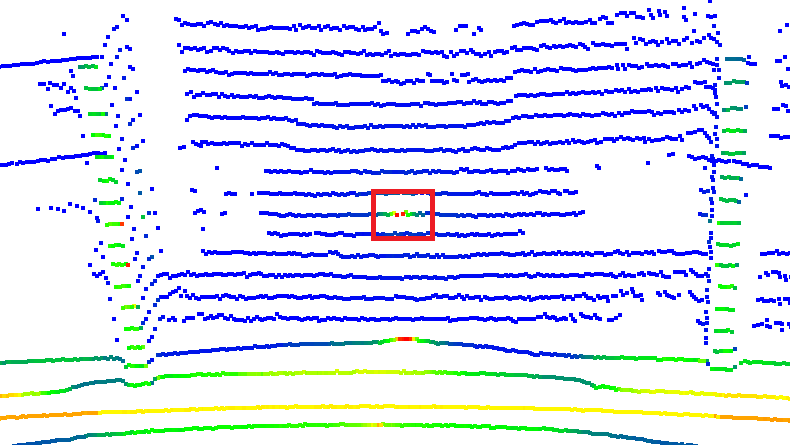}
\label{floortoceilingintensity}
}
\subfloat[Dual return]{
\includegraphics[width=4cm,height=2cm]{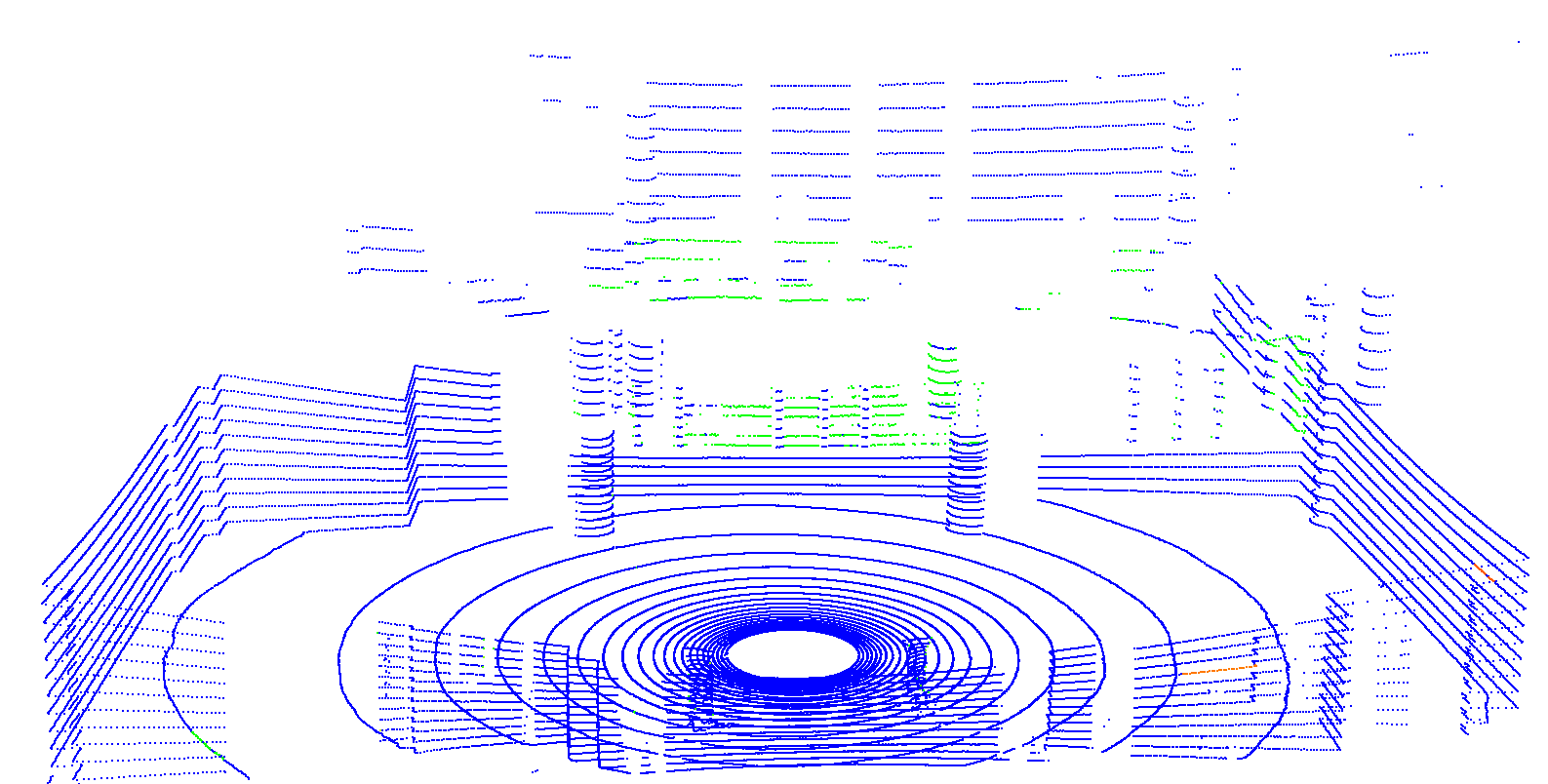}
\label{dualreturn}
}
\caption{Pointcloud examples}
\vspace{-0.2cm}
\end{figure}

The colors in 
\ref{floortoceilingintensity} represent the intensities of the points, a color closer to red means a stronger intensity. The red rectangles indicate the areas where intensity peaks exist. 


The point cloud shown in Fig. \ref{floortoceilingintensity}, is in a corridor with lots of floor to ceiling windows, which is No. 3 in the robot path in Fig. \ref{starcenterpath}. The background obstacles are far enough, so they do not have enough intensity. As a result, most of the glass can be observed, and the intensity peak is in the center of it.
We can also find that in the classroom scene Fig. \ref{dualreturn}, the robot is not high enough and no intensity peak can be found.
These examples show, that the intensity peak occurs when the beam is perpendicular to the reflective material.

\subsection{Dual Return Reflection Analysis}
To highlight the dual return analysis we get two point clouds in two different places and separate the last and the strongest point cloud. For the analysis, we manually choose the windows and remove (crop) points in front of the windows. Then we count all remaining points, the points just on the windows, the wrongly reflected points and the points of obstacles behind the windows.

The point cloud in Fig. \ref{dualreturn} is a combination of both the strongest return and the last return, where the points in the last return or in the intersection of last and strong return are colored as the blue ones, and the points in the strongest return but not in the last return are colored according to their intensity, from green, the least intensity, to red, the highest intensity. This point cloud is scanned in a classroom with some windows, and it marked as No. 1 in the robot path Fig. \ref{starcenterpath}. In this scene, there is no intensity peak (as the points on the window are all with low intensity, colored as green), because the robot is not high enough. As a result, the intensity peak method won't work here.

We define the set purely with points of glass to be $G$, with points of reflection to be $R$, with outdoor obstacles to be $O$, the set of normal, unreflected indoor points $I$, the set of last return points to be $l$ and strongest return to be $s$. 


First we consider the case where, measured from the glass, the outdoor obstacle is farther than the indoor obstacle: if $l \in O$, then $s \in G, R, O$; if $l \in R$, then $s \in G, R$; if $l \in G$, then $s \in G$. When $l \neq s$, we have $s \in G, R$ and $l \in O, R$.

Secondly, if, again measured from the glass, the outdoor obstacle is closer than the indoor obstacle: if $l \in O$, then $s \in G, O$; if $l \in R$, then $s \in G, R, O$; if $l \in G$, then $s \in G$. When $l \neq s$, we have $s \in G, O$ and $l \in O, R$.

Since we can use the phenomenon above to detect the glass, and we can have knowledge of the depth of the scenario and the outdoor obstacle through the scan, we may able to classify points in a scan into either $R, G, O$ or $I$. After we have the coefficients of the reflective plane fitted using $G$, we are able to mirror the reflection back to where the real objects are to achieve greater scan point utilization and actually also map areas not in the line of sight of the scanner.

\section{Methods}
\label{chap:methods}
According to the phenomenons and rules we concluded above, we design a pipeline to process the point cloud and detect reflection on a planar, rectangular and framed surface.
\begin{itemize}
\item Process Every Scan
    \begin{itemize}
        \item Process Velodyne packet and convert to organized point cloud
        \item Detect reflection using two approaches:
        \begin{itemize}
            \item Intensity Peak: find intensity peak and get a perpendicular infinite plane
            \item Dual Return: Choose nearer different strongest points and get an infinite plane
        \end{itemize}
        \item Find Boundaries: Get boundaries of reflective infinite plane based on the frame
        \item Classify Points and Mirror Reflected Points
    \end{itemize}
\item Integrate to SLAM Framework: Use filtered point cloud to do SLAM and get transform
\end{itemize}

\subsection{Process Vedolyne Data Packet and Convert to Organized Point Cloud}
 We use PCL\cite{Rusu_ICRA2011_PCL} to store, reorder and process the point cloud. We firstly separate the packets from Velodyne into strongest and last returns. These point clouds are unorganized. An organized point cloud can be considered as a 2D matrix. Velodyne HDL-32E has 32 rings, so we set the row to be 32. The step azimuth is 2.79 mrad and there are about 2251 points a ring, so the size of the organized point cloud is $32\times 2251$. We organize these point clouds according to the azimuth and channel id.

\subsection{Intensity Peak}
To find the intensity peak, we firstly choose the horizontal ring of the Velodyne from a total of 32 rings using the whole organized point cloud and traverse all points in the ring. We check if the intensity increases from a low threshold to a max threshold first and then decreases in the same way. Also, the distance of two adjacent points in the sequence should be less than a small threshold, because the points of the intensity peak in the horizontal ring on the glass are close. If there is a gap in the distance, the sequence is invalid or it should be ended at the gap. There may be multiple potential peaks, so we need to verify the points near the potential peak. We choose the points of the same degree in two up and two down rings and check the intensity on vertical is also in such order. If this is verified, we can store these points, and later fit a plane through those points. This Algorithm \ref{algorithmintensity} is described below.
\begin{algorithm}[htbp]
  \caption{Algorithm of Finding Intensity Peak}  
  \label{algorithmintensity}
  \begin{algorithmic}[0]  
    \Require  
      $cloud$: organized strongest point cloud of one scan
    \Ensure  
      Points of every Intensity peak $peak\ result$
    \State Filter the point cloud to horizontal plane of the sensor(z axis) save to $horizontal\ scan$;
    \For {each $point$ in $horizontal\ scan$ }
    {
        \State calculate $gap$ between this point and last;
        \If {$gap > threshold$}
            \If {the point is in a sequence increasing to the max intensity and decreasing sequence to next gap}
                \State Add the point to the sequence
            \EndIf
        \Else 
            \If {the sequence is valid}
                \State Store the peak to $potential\ peaks$
            \EndIf
        \EndIf
    }
    \EndFor
    \For {each $peak$ in $potential\ peaks$ }
    {
        \State Choose points from same azimuth degree (same column) in the upper and lower two rings
        \If {Verify the points is also an increasing then decreasing sequence}
            \State Store these points to $peak\ result$ 
        \EndIf
    }

    \EndFor
  \end{algorithmic}  
\end{algorithm}

We use random sample consensus (RANSAC)\cite{Fischler1981ransac} to fit the planes. We get the inliers of the plane as well as the plane parameters.

\subsection{Dual Return}
As described before in the dual return analysis, we can use the dual return method to find reflections. Firstly, we use both organized point clouds we separated from the dual return packet. In each ring, we check if the strongest and last point cloud of the same beam is the same. If they are the same, we keep them. If not, in that direction there exists glass. The closer point between the strongest and last point may be the one on the glass. We use RANSAC to fit the planes, which only works if the plane haves enough inliers. The Algorithm \ref{algorithmdual} is described below.
\begin{algorithm}[htbp]  
  \caption{Algorithm of Detect Reflection using Dual Return}
  \label{algorithmdual}
  \begin{algorithmic}[0]  
    \Require  
      $cloud$: organized strongest point cloud of one scan and organized last point cloud of one scan
    \Ensure  
      Degree of glass and the fitted plane of glass
    \For {each $point$ in each ring (row of cloud)}
    {
        \State Calculate the distance of the strongest and last point in the same degree
        \If {distance is the same}
            \State Add to $normal\ points$
        \Else
            \State Save the point with lower distance and in strongest point cloud to $glass\ points$
            \State Save rest points to $remain\ points$
            \State Save the rings and degree to $degree\ has\ glass$
        \EndIf
    }
    \EndFor
    \While {number of $glass\ points$ $>threshold$}
        \State Fit plane using RANSAC
        \If{plane has enough inliers}
            \State Save the fitted plane
        \EndIf
        \State Remove inliers from $glass\ points$
    \EndWhile
  \end{algorithmic}  
\end{algorithm}

\subsection{Find Boundary}

Through the two methods above, we can have some information including planes from intensity peak and planes from dual return and degrees that contain glass from dual return. Using this information we can estimate the boundaries of the glass. We have a hypothesis that the glass is installed in a metal frame or on a wall. So it should have a boundary in the form of a frame. Velodyne only has 32 rings vertically, so the upper and lower boundary may not be observed directly but the left and right boundary can be observed if there is a line of sight (LoS).
Since we have the planes, we search for the left and right points that are near the plane and in the direction of the glass.
Choose the nearest boundary for each plane as the left and right boundary. For the up and down boundary, check the angle that contains glass in each ring, choose the uppermost (lowermost) ring that contains glass as the upper (lower) boundary.

\subsection{Classify Points}
Now we have found some glass with boundary, so we need to classify all the points. We classify the points as inside objects $I$ , glass $G$, reflected points $R$ and obstacles behind glass $O$. The points exactly on the same plane within the boundary of the glass is $G$. The points in front of the glass plane is $I$ and the points behind glass consist of both $R$ and $O$. Since we detected the glass, it is simple to separate $I$ and $G$, but it is not easy to distinguish $R$ and $O$. Here we propose a three-step method to recognize whether a point is in $R$ or in $O$. We classify points using all detected glass planes and we ignore the plane if the number of outside points classified by that plane is too low.

The first and second step is to determine the points definitely belonging to $O$. In first step, we mirror the inside points $I$ against the glass plane. Outside points more far away than the mirrored points are considered as outside obstacles $O$. We then mirror the points behind glass back against the glass plane (to the inside) and trace the laser beams on the direction of mirrored points to check if there is another point further than it. If so, the origin point of this mirrored point is an outside obstacle. The procedure in step three is identical to step two, except we now have the points of outside obstacle, so we can determine the points in $R$ by tracing the laser beams on the direction of outside points and check if there is another point further than that in the same direction. 

\subsection{Reflection Back Mirroring}
Suppose the reflection point cloud is $R$, the parameters of the reflective plane is described in $P = \left(a, b, c, d\right)^T, d > 0$, where $N = \left(a, b, c\right)^T$ is the normalized normal vector of this plane, and the mirrored point cloud is $M$. According to the Householder transformation \cite{householder1958unitary}, we can mirror the reflection point back by
\begin{align*}M = R \cdot \left[\begin{matrix} I-2\cdot N\cdot N^T & -2d\\ 0_{1\times 3} & 1 \end{matrix}\right]\end{align*}

\subsection{Integration to SLAM Framework}
To understand how our method can improve the SLAM quality, we integrated it to a SLAM framework. For 3D SLAM, we use HDL Graph SLAM\cite{Koide2019slam}, which is a state of art SLAM method and robust in different environments. We add inside points, mirrored points and glass points and outside obstacles points to the SLAM point cloud. We modified HDL Graph SLAM to SLAM with the SLAM point cloud and get the transform. We save the position of detected glass planes with the transform in the map frame. When we can not find the glass plane by previous methods, we find the saved glass plane based on the current pose to classify the point. After SLAM we have a good quality map without glass interference and with points classified to different labels. We can also use the transform of the SLAM to build a map of glass for obstacle avoidance and further use. 

\section{Experiments and Discussion}
\label{chap:results}
We record the experiment data by driving a robot in the STAR Center second floor. The map and the path of the robot are shown in Fig. \ref{starcenterpath}. The experiment data includes the raw Velodyne packet and the original dual point cloud with 10Hz, respectively. We use an Intel i7-8750H laptop to run the experiments. Running with one CPU core, the average process time of one point cloud is 29ms, so the the algorithm can run in real time. 

\begin{figure}[htbp]
\centering
\subfloat[Classroom]{
\includegraphics[width=2.6cm,height=1.8cm]{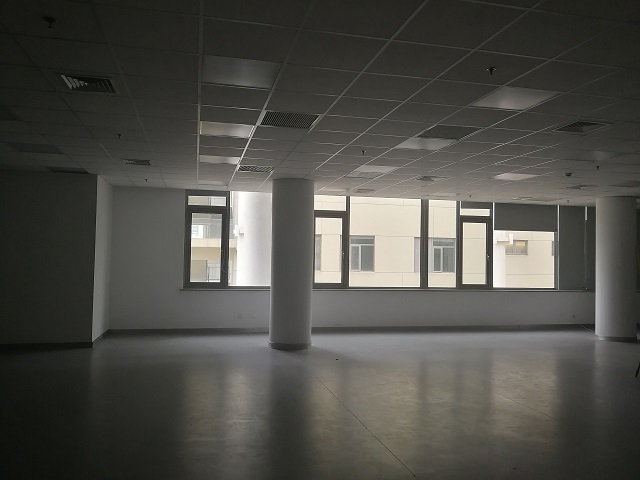}
\label{classroom}
}
\subfloat[Glass railing]{
\includegraphics[width=2.6cm,height=1.8cm]{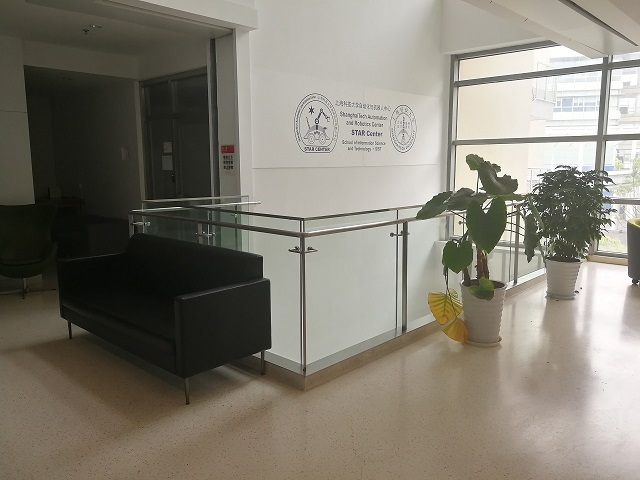}
\label{glassrailing}
}
\subfloat[Floor to ceiling glass]{
\includegraphics[width=2.6cm,height=1.8cm]{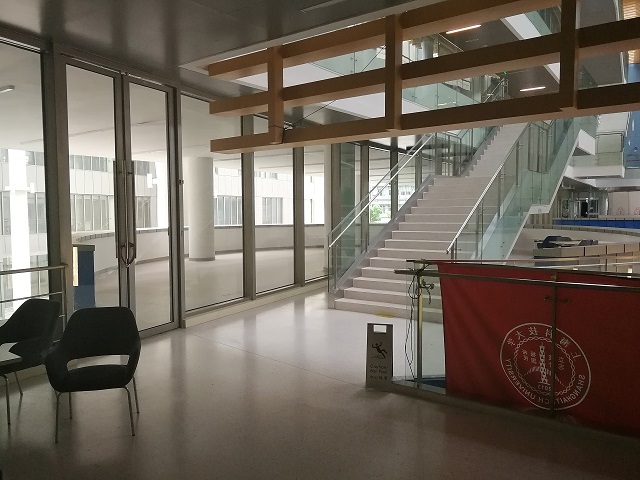}
\label{floortoceilingwindows}
}
\caption{Photos of three experiment scenes}
\end{figure}

The photo of three scenes in the map is shown in the following. Fig. \ref{classroom} is a classroom with some windows, Fig. \ref{glassrailing} is a corridor with glass railing, Fig. \ref{floortoceilingwindows} is a corridor with lots of floor to ceiling glass.

\subsection{PointCloud Classification Evaluation}

We evaluated the dual return method in the classroom scenes. We use two types of ground truth points: Velodyne and FARO. For the inside points and glass points we collect and average 300 point clouds from the sensor. For ground truth collection the windows are covered with electrostatic cling paper and their points are manually classified, using the FARO data as an aid. Since the sensor is not moved during and between ground truth collection and experiment data collection, the same points will be sampled. 

The reflected points and outside ground truth points cannot be collected this way, as we have no way to get the ground truth information if a point was reflected or passed through the window. So we use a professional 3D Scanner (FARO) for this data, scanning inside the room (for reflected points) and outside (for outside points). All ground truth points are first manually aligned and then registered with ICP for maximum accuracy. 

\begin{figure}[htbp]
\centering
\vspace{0.2cm}
\subfloat[Classroom]{
\includegraphics[width=8.5cm]{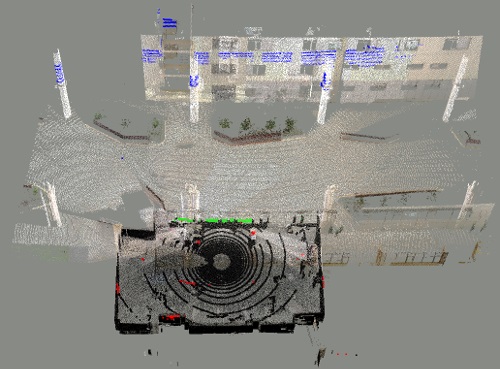}
\label{classroom_detect}
}

\caption{Detection Result}
\end{figure}

In the experiment we remove the cling paper and capture 340 pointclouds with the Velodyne sensor. Fig. \ref{classroom_detect} shows the ground truth FARO point cloud (subsampled for visualization) and one experiment point cloud. The black points are the real objects inside, the green points are the points of glass, the blue points are the points of obstacles outside and the red points are the reflections mirrored back against the glass. 

Given the known ground truth position for all points we can now evaluate our classification result by comparing the estimated position of that point with the ground truth position. For that we assume a maximum error of 0.15m for a correct classification. Additionally we can quantify the range error for the different point classes, which is increased due to the laser's interaction with the glass. The results are shown in Table~\ref{Classificationresulttable}.



\begin{table}[htbp]
\centering
\caption{PointCloud Classification Evaluation}
\begin{tabular}{|l|c|c|c|c|}
\hline
    \textbf{Classified points} & \textbf{\% \textless0.1m} & \textbf{\% \textless0.15m} & \textbf{Mean} & \textbf{Stdev}\\\hline
    Reflection  & 73.9 & 92.8 & 0.106 & 0.371\\ \hline
    Glass  & 96.3 & 99.9 & 0.0534 & 0.0276\\ \hline  
    Outside Obstacle & 92.6 & 97.5 & 0.0559 & 0.0924\\ \hline  
    Inside Obstacle & 99.8 & 99.9 & 0.00307 & 0.138\\ \hline
\end{tabular}
\label{Classificationresulttable}

\end{table}

\subsection{Glass Plane Evaluation}
We use two methods to get the glass plane. At classroom scenes we use the dual return method. At the glass railing and floor to ceiling glass scenes we use the intensity peak method. For each scene, we collect about 300 pointclouds in the same place. We also use the stickers to get the ground truth pointclouds of glass. We detect planes in each pointcloud and calculate the distance from the ground truth glass points to the plane. So we can get the RMS error on average, the result shows in Table~\ref{planeresulttable}. We also detect plane on ground truth plane point and calculate the average angular difference in degree. 

\begin{table}[htbp]
\centering
\caption{Plane Fitting Result}
\begin{tabular}{|l|r|r|r|}
\hline
    \textbf{Average Error} & \textbf{RMS} & \textbf{\% RMS\textless0.08} & \textbf{Angular in \degree}\\  \hline
    Classroom  & 0.0577 & 96.7& 2.87\\ \hline
    Glass Railing  & 0.0232 & 100 & 3.03\\ \hline
    Floor to ceiling glass  & 0.0716 & 86.4& 10.04\\ \hline
\end{tabular}
\label{planeresulttable}
\vspace{-0.5cm}
\end{table}

\subsection{SLAM Experiment}

We integrate the methods to HDL Graph SLAM and run the mapping twice with all the data recorded, the first time using the original dual point cloud and then using the SLAM point cloud described above. For comparison and statistics, we label the points in the original SLAM manually.

\begin{figure}[htbp]
\centering
\subfloat[SLAM on original point clouds (color labeled manually)]{
\includegraphics[width=8.5cm,height=3cm]{Experiments/Figures/originslam.png}
\label{originslam}
}\\
\subfloat[SLAM on classified point clouds]{
\includegraphics[width=8.5cm,height=3cm]{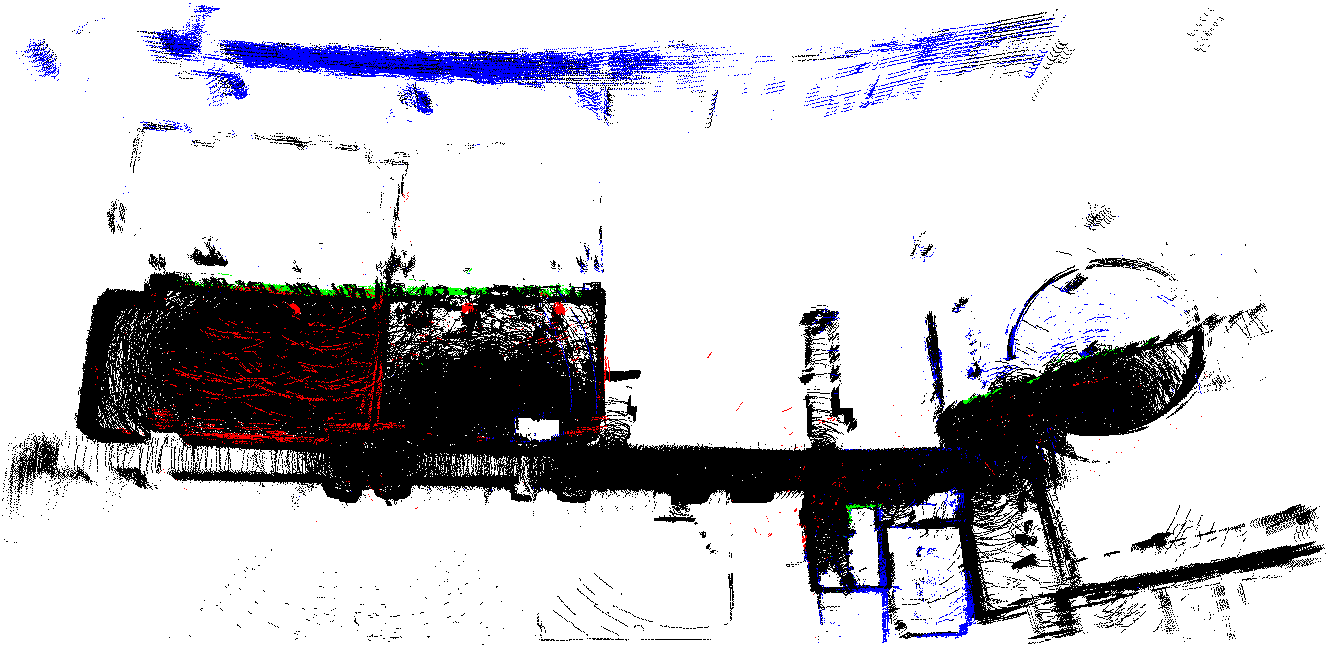}
\label{filteredslam}
}
\caption{SLAM Result}
\label{slamresult}
\vspace{-0.3cm}
\end{figure}

\begin{table}[htbp]
\centering
\caption{SLAM Experiment Result}
\begin{tabular}{|l|r|r|}
\hline
    \textbf{Original SLAM} & \textbf{\#Points} & \textbf{Percent} \\  \hline
    Total points  & 2,030,898 &  \\ \hline
    Reflection points  & 54,396 & 2.68\\ \hline
    Glass points  & 19,769 & 0.97\\ \hline  
    Outside Obstacle points  & 164,336 & 8.09\\ \hline  
    Inside Obstacle points & 1,792,397 & 88.26 \\ \hline \hline
    \textbf{Classified SLAM} & \textbf{\#Points} & \textbf{Percent} \\  \hline
    Total points  & 1,765,835 &  \\ \hline
    Glass points& 12,978 &  0.73\\ \hline
    Mirrored points  & 13,908 & 0.79 \\ \hline
    Unclassified reflection points  & 2,799 & 0.16 \\ \hline
    Inside Obstacle points & 1,672,622 & 94.72 \\ \hline
    Outside obstacle points & 63,528 & 3.60 \\ \hline
    --Correct Marked Outside obstacle points & 47,601 & 2.70 \\ \hline
\end{tabular}
\label{resulttable}
\end{table}

Fig. \ref{slamresult} shows the two SLAM results with the original point cloud and with the classified point cloud. Table \ref{resulttable} shows the counts of points in different SLAM results. From the result, we can find that the original SLAM has far more reflection points than classified SLAM, which shows that our method is effective. Most of the glass points are detected and classified. A lot of reflection points have been mirrored back in the classified result and the reflections of the ceiling are also detected successfully, which makes it possible to have the knowledge of the room height using just one horizontal scan, which normally cannot see the ceiling. However, there are still a few unfiltered reflection points in the classified results. This is because we may not observe the full glass and can only find the nearest boundary of the glass, so we can not remove the reflection outside the boundary of the glass. For the obstacles in the classified SLAM, our method marks points behind the glass railing as outside obstacles. Some outside obstacles are not marked because the points have the same distance in strongest and last point cloud, and thus they can not be classified. Most importantly, the mirrored reflection points can successfully map objects not in the line of sight. For example, the back side of the three pillars and the ceiling in the classroom can not be observed in the original SLAM.

\subsection{Mapping with Reflection}
\begin{figure}[htbp]
\includegraphics[width=\linewidth]{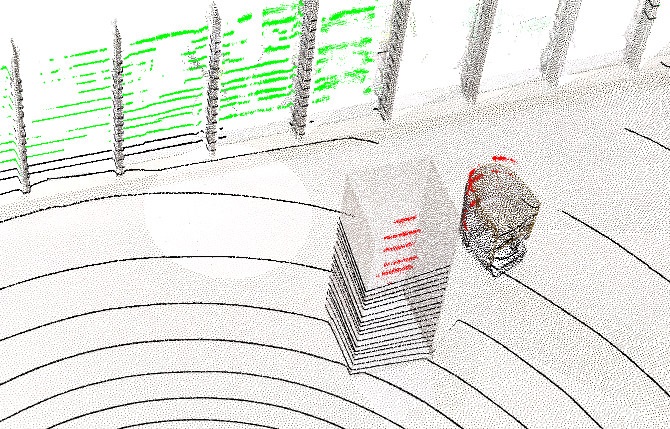}
\caption{Mapping result with detected glass (green) and reflection (red)}
\label{reflectionmapping}
\end{figure}

As mentioned in previous experiment, the reflected points can be used to build the map. One of the advantages is to map the back side of the objects that are normally blocked. Fig. \ref{reflectionmapping} has demonstrated the points on the back sides of the objects, marked as red. The reflective plane is marked as green and the points on normal surfaces are marked as black. The points with smaller sizes are the ground truth data from the FARO scanner. The reflected points match the object both in the point cloud from Lidar scanner and from the FARO scanner with a small misalignment. This misalignment is caused by the small noise of estimating the reflective planes, which leads to a bigger error in the reflected points.

\section{Conclusions}
\label{chap:conclusions}

We developed a method for reflection detection and utilization using a dual return Lidar, which is useful for indoor SLAM. This method is able to detect different reflective materials, such as glass railing, glass door and floor-to-ceiling windows. The intensity peak approach can successfully detect the glass. Using the dual return method, we can successfully find the glass plane and the range of scan-angles that contains glass. Classification of the points is achieved with the boundary of glass fitted through the detected planes. This classification also enables us to mirror the reflected points back to achieve mapping of objects behind the line of sight and out of field of view. 

Our experiments show that we are able to correctly classify the points, estimate the planes of the glass, create pointcloud maps with according classifications and, utilizing the reflection, map non-line-of-sight objects, effecitvely achieving "mapping around the corner". 
%

As a future work we aim to integrate the mirror detection itself into a SLAM framework, such that a more robust and accurate mirror and window detection is possible, which then also enables even more accurate point classification and reflection utilization. Furthermore, we are considering the detected glass planes as features for SLAM.

\IEEEtriggeratref{2}

{\small
\bibliographystyle{IEEEtran}
\bibliography{egbib}
}

\end{document}